\documentclass[runningheads]{llncs}
\usepackage{graphicx}
\usepackage{amsmath}
\usepackage{amssymb}
\begin{document}
\title{TIFu: Tri-directional Implicit Function for High-Fidelity 3D Character Reconstruction}
\titlerunning{TIFu}
%
\author{Byoungsung Lim\inst{1} \and
Seong-Whan Lee\inst{1}}
%
%
\institute{Korea University, Seoul 02841, South Korea}
\maketitle              
\begin{abstract}
Recent advances in implicit function-based approaches have shown promising results in 3D human reconstruction from a single RGB image. However, these methods are not sufficient to extend to more general cases, often generating dragged or disconnected body parts, particularly for animated characters. We argue that these limitations stem from the use of the existing point-level 3D shape representation, which lacks holistic 3D context understanding. Voxel-based reconstruction methods are more suitable for capturing the entire 3D space at once, however, these methods are not practical for high-resolution reconstructions due to their excessive memory usage. To address these challenges, we introduce Tri-directional Implicit Function (TIFu), which is a vector-level representation that increases global 3D consistencies while significantly reducing memory usage compared to voxel representations. We also introduce a new algorithm in 3D reconstruction at an arbitrary resolution by aggregating vectors along three orthogonal axes, resolving inherent problems with regressing fixed dimension of vectors. Our approach achieves state-of-the-art performances in both our self-curated character dataset and the benchmark 3D human dataset. We provide both quantitative and qualitative analyses to support our findings.

\end{abstract}

\section{Introduction}
\label{sec:intro}
Creating 3D animation characters is a challenging task that demands significant time and effort, especially in the gaming and movie industries. With the rise of new technologies such as augmented reality (AR), virtual reality (VR), and the Metaverse, there is growing interest in replicating 3D human models using various approaches \cite{R4,R9,M1,M4,E21,E22}.
Recent advances \cite{R8,R10,R11,R12} in 3D deep learning have enabled the reconstruction of highly detailed 3D human models from a single image, which is a significant breakthrough in the field. However, there has been relatively little exploration or analysis of the use of deep learning models for reconstructing 3D animated characters, despite their potential to revolutionize the field.

\begin{figure*}
  \centering
  \includegraphics[width=\linewidth]{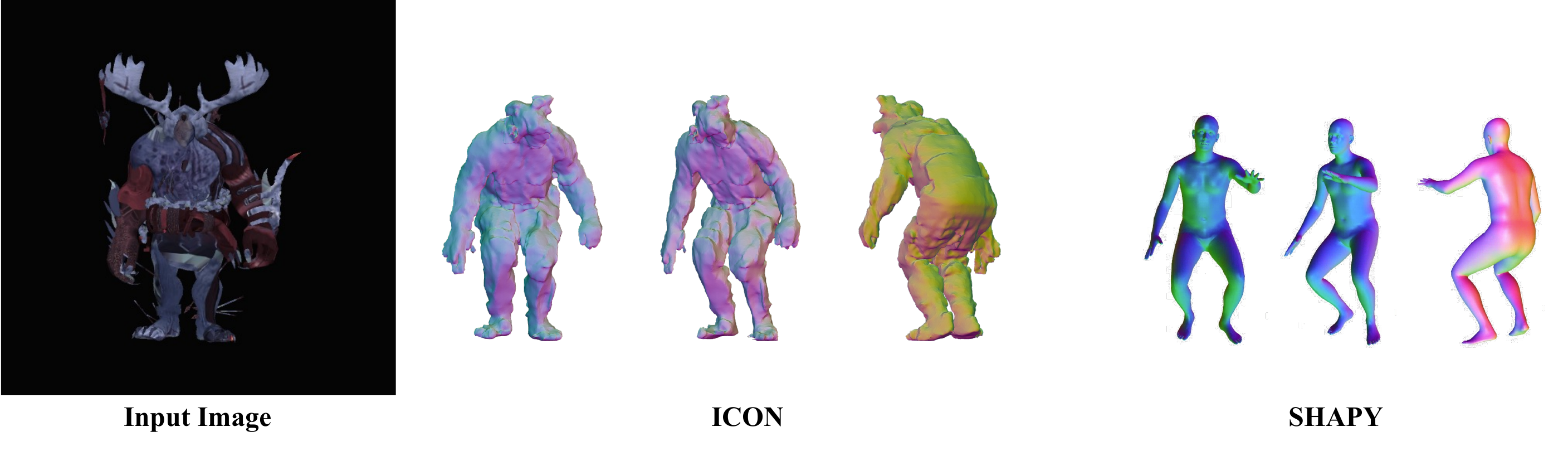}

  \caption{\textbf{Qualitative results from template-based methods for character reconstruction.} Template-based methods \cite{R10,E20} fall short when it comes to reconstructing characters with significant geometrical differences from humans.} 
  \label{fig:template}
  \vspace{-2em}
\end{figure*}

Our goal is to reconstruct high-fidelity 3D character mesh from a single image with sufficient generalization ability to extend into in-the-wild environments. However, existing 3D human reconstruction methods \cite{F3,F4,F6} face several challenges when applied to 3D animated characters.

First, animated characters exhibit much greater shape variation than humans, making it difficult to achieve accurate reconstructions using existing 3D human reconstruction methods. State-of-the-art single-view 3D human reconstruction methods rely on Pixel-aligned Implicit Function (PIFu) \cite{F3} representation, and we found that PIFu-based methods are only effective when there are large geometrical similarities between the training and test data distribution. When inferring out-of-distribution images, PIFu-based methods tend to generate dragged geometry or spiked mesh surfaces along the depth from the camera view. Second, while there may be many approaches to improve the reconstruction quality of 3D human models, such as leveraging human templates \cite{P1}, the large variations in 3D character shapes make it extremely challenging to use existing templates or create new ones that can represent such variations with the fixed number of parameters. Lastly, voxel representations \cite{V1,V2,V3} propagated in a fully convolutional manner are typically more robust towards generating globally consistent 3D geometries. However, the memory-intensive nature of this representation is not suitable for detailed 3D mesh reconstruction.

To address these challenges, we propose an end-to-end multi-level framework with a new Tri-directional Implicit Function (TIFu) representation for 3D character reconstruction. TIFu is a vector-level representation that combines the benefits of pixel-level feature learning and voxel representations. It estimates a single line of voxels, which we refer to as a \textit{vector} throughout this paper, that can be inferred parallel to three orthogonal Cartesian axes. In contrast to existing PIFu-based methods that locally align pixel-level information to the 3D point's binary occupancy representing inside and outside the mesh, TIFu learns to map from latent shape encodings to vector-level occupancies in three orthogonal directions. This approach is more memory-efficient than using voxel representations and has shown improved performance in inferring unknown shapes compared to inferring occupancies per point.

We base our multi-level architecture on \cite{F4}, which highlights the importance of incorporating high-resolution image features to increase model expressiveness in capturing fine surface details. Although this approach has improved the quality of reconstruction enough to recognize facial features and clothing folds, it often fails to reproduce natural-looking geometry when seen from another angle compared to the camera view. Our method differs in that we incorporate all geometric representations from both coarse and fine levels, which allows us to seamlessly fuse the globally consistent low-resolution geometry and emboss high-resolution surface details on top, enabling the retention of pixel-level visual cues strongly guided by the input image.
Moreover, inferring a fixed dimension of vectors would be insufficient to recover the target 3D space at an arbitrary resolution, unlike existing PIFu-based methods. However, we have developed a new algorithm that allows us to aggregate tri-directional shape inference, compensating for each other to complete reconstruction at a desirable resolution.

The main contributions in this work consists of:
\begin{itemize}
    \item an end-to-end framework with a novel vector-level 3D representation to tackle 3D reconstruction of character geometries for the first time.
    
    \item a method to alleviate the limitation of inferring fixed size of vectors while incorporating multi-level geometric representations.

\end{itemize}
\section{Related Work}
\label{sec:related}

We find our work as the extreme extent of 3D human reconstruction tasks where we will mainly review single-view 3D human reconstruction methods. 

\noindent\textbf{Template-based Methods.}
Since multiple 3D geometric configurations can be projected into the same 2D image from a camera view \cite{E11}, we often employ domain-specific templates \cite{P1,P2,P3}, consists of parameters controlling shapes and poses, to effectively overcome such ambiguity. However, templates lack in ability to capture loose clothes or hairstyles beyond the restricted low-dimensional subspace of the underlying linear statistical model \cite{E4}. Researchers also proposed methods \cite{T2,T22} to further deform a reconstructed mesh obtained from the template to better fit the target geometry. Still, these methods cannot extend to shapes with large differences from the template as shown in Fig. \ref{fig:template}.

\noindent\textbf{Template-free Methods.}
Subsequently, various approaches have been developed to reconstruct 3D geometry in a free-form manner without relying on templates. These methods have significantly improved expressiveness in handling unknown shapes, but they often produce falsified mesh artifacts due to inaccurate shape understandings. Template-free methods primarily differ in their output shape representation, which includes voxel-based, implicit function-based, and neural radiance fields (NeRF)-based approaches \cite{N1}.

Voxel-based methods learn to estimate a predefined resolution of 3D voxels where each voxel represents the occupancy of 3D shapes in space. Varol \textit{et al.} \cite{V1} proposes BodyNet that incorporates pose estimation and body part segmentation to estimate the volumetric shape of a human body. Zheng \textit{et al.} \cite{V3} introduces DeepHuman and further improves the details of reconstruction by applying additional information guided by the surface normals. Since these methods require memory intensive training and the spatial resolution is mostly limited to \(128^3\) \cite{E8}, the resulting shapes often lack fine surface details.

Implicit function-based approaches \cite{C1,F10,F11,F12} offer a significant improvement over voxel-based methods as they can continuously map between a point in 3D space and its 3D shape representation, discriminating between inside and outside the mesh. This capability provides a memory-efficient way to reconstruct complex 3D geometries at an arbitrary resolution without explicitly storing huge voxels. Pioneering the use of this approach, Pixel-aligned Implicit Function (PIFu) \cite{F3} allows single-view 3D human reconstruction by mapping pixel-level information to spatially aligned 3D shape representations. Building on this work, PIFuHD \cite{F4} can capture fine surface details such as facial features or clothing folds by leveraging feature resolution up to 1K, while Geo-PIFu \cite{F6} combines additional 3D voxel features with pixel-aligned features to better capture global 3D context and achieve more natural reconstruction among adjoining body parts. Hybrid models \cite{F1,T3,R3} like ARCH \cite{F2} and ARCH++ \cite{F9} combine existing templates with implicit modeling techniques by leveraging SMPL \cite{P1} to define a deformation field that enables 3D modeling in canonical space and animating reconstructed human mesh. Despite their impressive results for a specific domain, these methods may fail to generate consistent results when there is a variant data distribution.

NeRF-based techniques \cite{R2} are gaining popularity for creating photorealistic renderings of novel viewpoints from a single or few images. The primary objective of these approaches is to recover colors at previously unseen directions, but they can also be used for 3D shape reconstruction by utilizing the density field learned from multiple views. However, NeRF-based methods optimized for single-image-based 3D reconstruction, such as pixelNeRF \cite{N2}, have limitations in producing high-quality 3D models. This is because NeRF is designed to encode a hazy feature space that best reconstructs 2D renderings without explicit 3D geometric understanding, resulting in smooth surfaces lacking in fine details. Additionally, these techniques typically rely on multiple images to model accurate 3D geometries, further restricting their effectiveness for single-view reconstruction.
\section{Method}
\label{sec:method}
\begin{figure*}
  \centering
  \includegraphics[width=\linewidth]
  {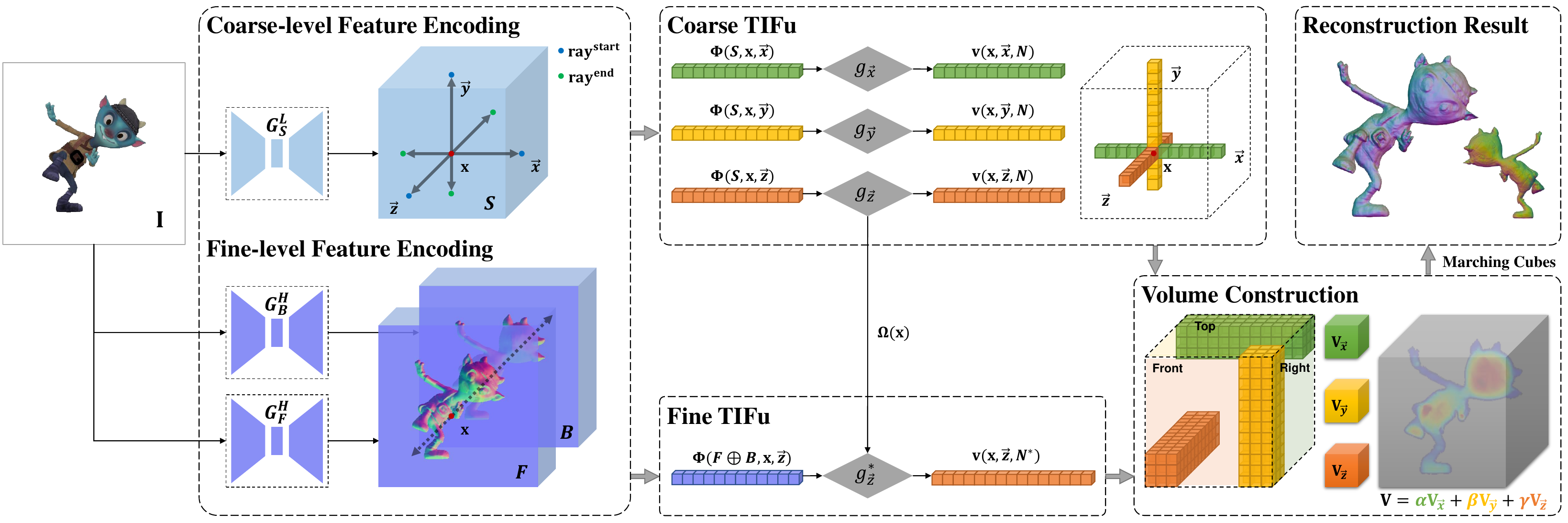}

  \caption{\textbf{Overview of our 3D mesh reconstruction.} Our approach involves constructing a 3D space using vectors along three orthogonal axes in a coarse-to-fine manner. Coarse-level module estimates vector-level 3D representations along tri-directional rays based on a given query point. We then refine our coarse-level vectors along depth by attending to high-resolution visual cues. The final 3D volume is constructed by stacking densely acquired vectors and aggregating the resulting three separate volumes. We obtain the 3D mesh by applying Marching Cubes to the final 3D volume.}
  \label{fig:overview}
  \vspace{-1.5em}
\end{figure*}

In this section, we describe 1) the background on implicit 3D modeling (Sec. \ref{sec:background}), 2) the tri-directional implicit function (TIFu) representation (Sec. \ref{sec:tifu}), and 3) the process of coarse-to-fine vector-level 3D shape inference (Sec. \ref{sec:coarse-to-fine}). 

\subsection{Background on Implicit 3D Shape Modeling}
\label{sec:background}

The basic principle underlying learning of an implicit function \(f\) is to estimate a 3D shape representation \(s\) based on a given query point \(\mathbf{x}\) in 3D space:
\begin{equation}
    f(\mathbf{x}) = s: \mathbf{x} \in \mathbb{R}^3, s \in \mathbb{R}.
\end{equation}
One commonly used 3D shape representation is the binary occupancy function, which determines whether a point in 3D space is inside the watertight mesh or not. Another frequently utilized representation is the Signed Distance Function (SDF) \cite{E8}, which calculates the distance between a point and the nearest mesh surface and assigns a sign based on whether the point is inside (negative) or outside (positive) the mesh.

Pixel-aligned Implicit Function (PIFu) \cite{F3} is a widely used method for implicit function learning and has become a technical foundation for many recent single-view 3D human reconstruction methods. PIFu predicts the binary occupancy value for any given point in 3D space by aligning pixels with the underlying 3D geometry. To achieve this, PIFu first extracts an image feature embedding based on the 2D location where a 3D point is projected onto the input image. This embedding is then concatenated with the point's depth-wise distance from the camera and propagated through a Multilayer Perceptron (MLP) to classify the occupancy of the point. Therefore, during training, the process involves sampling 3D points and computing the occupancy loss at these sampled locations without generating 3D meshes. During inference, the target 3D space is uniformly sampled to infer the dense 3D occupancy volume, and the final mesh is constructed using the Marching Cubes \cite{E1}.

\subsection{Tri-directional Implicit Function}
\label{sec:tifu}

Our goal is to reconstruct highly variant 3D character shapes from a single image while preserving global consistencies and fine details. To achieve this, we introduce a memory-efficient and robust 3D shape representation called Tri-directional Implicit Function (TIFu), which combines implicit and voxel representations. In contrast to previous approaches that predict occupancy values for uniformly sampled 3D points or entire discretized voxels, we estimate a set of occupancies for grid points along a ray passing through a query point and parallel to one of the three Cartesian axes, which we refer to as \textit{vector}. By doing so, we can infer 3D shapes by each vector space, which captures the overall geometrical context more effectively than estimating point-level 3D representations that fail to consider neighboring point occupancies or understand the correlation required for natural-looking and plausible 3D geometry. Thus, the objective of TIFu is to model a function, \(\mathbf{v}(\mathbf{x})\), that predicts the binary occupancies of 3D points uniformly sampled from a minimum bound to a maximum bound based on a query axis \(\pi \in \{\vec{x}, \vec{y}, \vec{z}\}\) and a query point \(\mathbf{x}\) in 3D space:
\begin{equation}
    \mathbf{v}(\mathbf{x}, \pi, N) = \begin{bmatrix}
s_1, s_2, \hdots, s_N
\end{bmatrix} : s_i = f(\mathbf{x}_{i}) \in \mathbb{R},
\end{equation}
\begin{equation}
    \mathbf{x}_{i} = \text{ray}^{\text{start}}(\mathbf{x}, \pi) + \frac{i}{N} (\text{ray}^{\text{end}}(\mathbf{x}, \pi) - \text{ray}^{\text{start}}(\mathbf{x}, \pi)),
    \label{eq:x_i}
\end{equation}
where \(N\) denotes the number of evenly-spaced grid points \(\mathbf{x}_{i}\) along the ray. We use the functions \(\text{ray}^{\text{start}}\) and \(\text{ray}^{\text{end}}\) to calculate the start and end positions in 3D space for uniform grid points sampling along a query axis.
TIFu models the function \(\mathbf{v}\) using a neural network architecture that is trained end-to-end. Firstly, a cube-shaped 3D voxel features \(S\) is generated from an input image, which is spatially aligned with the target 3D volume. Next, a latent feature vector is extracted based on the relative location of a query point inside the feature space and the direction parallel to a query axis, in order to estimate the corresponding occupancy vector:
\begin{equation}
    \mathbf{v}(\mathbf{x}, \pi, N) = g_\pi(\Phi(S, \mathbf{x}, \pi)),
\end{equation}
where \(\Phi\) denotes the feature extractor that retrieves ray-aligned feature vector given a query point and an axis. To estimate the occupancy vector, we employ Multilayer Perceptrons (MLPs) for the functions \(g_\pi\).\\

\subsection{Coarse-to-Fine TIFu}
\label{sec:coarse-to-fine}

Vector-level 3D representation is more memory-efficient than voxel representations, but sustaining high-resolution features all the way to capture fine surface details can still limit model complexity in practical applications. To address this challenge, we propose a multi-level reconstruction framework that uses a coarse-level module to generate consistent geometries from three orthogonal directions and a fine-level module to capture detailed surface curvatures, such as facial features or clothing folds. This approach enables us to balance memory usage and model complexity, resulting in more accurate 3D models.

In the coarse-level module, we use a mapping function to transform latent voxel features into the corresponding vector space in three orthogonal directions, as described in Sec. \ref{sec:tifu}. To achieve high-fidelity reconstruction of a cube-shaped dense 3D occupancy volume, we divide the volume into vectors at a desirable resolution. For each face of the cube, we sample an arbitrary number of grid points inside the square face and infer vectors orthogonal to the face, with fixed vector dimensions. We then use linear interpolation to elongate each vector to match the desired resolution and integrate vectors inferred along each axis with matching dimensions. Our design choice of inferring vectors along three orthogonal axes is critical for learning coherent features and generating consistent geometries in all three directions, while also facilitating the ability to scale up the fidelity of the reconstruction.

As previously noted in a related study \cite{F4}, the model's ability to capture fine details is constrained by the resolution of the features. To address this issue, we propose to inject higher resolution surface normal features into the estimation of high-resolution vectors along the depth axis. We extract the high-resolution shape encoding in a pixel-aligned manner from the concatenated frontside and backside normal features. We then concatenate this encoding with the intermediate features extracted from the coarse-level MLP along the depth axis to infer finer-scale vectors that attend to visual cues at a higher resolution than our coarse-level reconstructions:
\begin{equation}
    \mathbf{v}(\mathbf{x}, \pi, N^*) = g_{\vec{z}}^*(\Phi(F \oplus B, \mathbf{x}, \vec{z}) \oplus \Omega(\mathbf{x})),
\end{equation}
where \(N^*\) denotes the length of high-resolution vectors and \(g_{\vec{z}}^*\) denotes the fine-level MLP. \(F\) and \(B\) are the frontside and backside surface normal features encoded from an input image, and \(\Omega(\mathbf{x})\) is the extracted intermediate features from the coarse-level MLP where \(\oplus\) denotes concatenation. To achieve complete reconstruction during inference, we first grid-sample query points from the front, top, and right faces at a desired resolution. We infer vectors for each query point, with vectors along the \(z\) axis inferred from the fine-level module and the \(x\) and \(y\) axes inferred from the coarse-level module. These vectors are linearly interpolated to match their dimensions and then stacked to construct dense 3D volumes. Finally, we weight-sum the volumes to obtain the final 3D volume \(\mathbf{V}\) where \(\alpha + \beta + \gamma = 1\):
\begin{equation}
    \mathbf{V} = \alpha \mathbf{V}_{\vec{x}} + \beta \mathbf{V}_{\vec{y}} + \gamma \mathbf{V}_{\vec{z}},
\end{equation}

\subsection{Loss Functions}
\label{sec:loss}
The proposed method encounters a significant class imbalance problem as most of the elements in the vector space are empty, particularly around thin body parts such as arms or hair. To address this issue, we utilize a Binary Cross Entropy (BCE) loss with an adaptive weighting mask \(\mathcal{M}\). To identify comparably thin body parts, we measure the distance from the first intersection point to the last intersection point by casting three orthogonal rays from each vector element when obtaining the ground truth vectors. We then consider the minimum distance among the three axes to account for the thinness at each location. Thus, the general loss term along each query axis, given a set of query points \(P,\) is defined as follows:
\begin{align}
\begin{split}
    \mathcal{L}_\pi &= \sum_{\mathbf{x} \in \mathcal{P}} \sum_{i=1}^{N} \mathcal{M}(\mathbf{x}_{i}) \bar{\mathbf{v}}_{i} \log{\mathbf{v}_{i}}\\
    &+ (1 - \bar{\mathbf{v}}_{i})\log{( 1 - \mathbf{v}_{i})},
\end{split}
\end{align}
\begin{equation}
    \mathcal{M}(\mathbf{x}_{i}) = 1 + f(\mathbf{x}_{i}) \times \frac
    { \delta } 
    { \displaystyle\min_{\forall\pi} {d(\mathbf{x}_i)}},
    \label{eq:w}
\end{equation}
where \(\bar{\mathbf{v}}_{i}\) and \({\mathbf{v}}_{i}\) denote the \(i^{\text{th}}\) element of the ground truth and the predicted vector. We apply a weighting mask to the inside points only that increases as the minimum distance \(d(\mathbf{x}_i)\) between the first and last intersection points from tri-directional ray casting gets lower. The scaling factor for this weighting mask is controlled by the hyperparameter \(\delta\). Therefore, the overall loss \(\mathcal{L}_\text{total}\) for the coarse-to-fine modules along the three orthogonal axes is computed as follows where \(\mathcal{L}_{\vec{z}}^{*}\) denotes the fine-level reconstruction loss:
\begin{equation}
    \mathcal{L}_\text{total} = 
    \alpha \mathcal{L}_{\vec{x}} 
    + \beta \mathcal{L}_{\vec{y}} 
    + \gamma \mathcal{L}_{\vec{z}} 
    + \mathcal{L}_{\vec{z}}^{*}.
\end{equation}

\section{Experiment}
\label{sec:experiment}
\subsection{Implementation Details}

Our framework consists of two types of modules: an encoder based on Convolutional Neural Networks (CNN) and a decoder based on Multilayer Perceptron (MLP). The low-resolution encoder architecture, denoted by \(G^L\), includes 2D convolutional downsamplings with 3D convolutional upsamplings in a U-net architecture, while the high-resolution encoder architecture, denoted by \(G^H\), only includes 2D convolutions. We employ four stacks of the \(G^L\) feature block for \(G^L_S\) and one stack of the \(G^H\) feature block for both \(G^H_F\) and \(G^H_B\). We first downsample an input image \(\mathbf{I}\) with dimensions of \(512 \times 512 \times 3 \) to \(128 \times 128 \times 128\) and and use it to encode the coarse-level 3D voxel features \(S\). We sustain the resolution for the fine-level features at \(512 \times 512 \times 32\). 
Specifically, \(g_\pi\) inputs ray-aligned features \(\Phi(S, \mathbf{x}, \pi) \in \mathbb{R}^{128} \) to infer \(\mathbf{v}(\mathbf{x}, \pi, N)\) where \(N=128\). 
Similarly, \(g_{\vec{z}}^*\) inputs \(\Phi(F \oplus B, \mathbf{x}, \vec{z}) \in \mathbb{R}^{32 + 32} \) concatenated with the intermediate features \(\Omega(\mathbf{x}) \in \mathbb{R}^{512} \) to infer \(\mathbf{v}(\mathbf{x}, \vec{z}, N^*)\) where \(N^*=512\). 
We jointly optimize the first three channels of \(F\) and \(B\) to predict the surface normal maps of the front side and back side of the mesh with an L1 loss function. Therefore, these high-resolution features represent not only surface normals but also unknown image features in the rest of the channels, such as relative depth information among adjacent body parts, which are optimized from our reconstruction loss.
To construct our final 3D volume \(\mathbf{V}\), we aggregate each axis' volume by \((\alpha, \beta, \gamma)=(\frac{1}{7},\frac{2}{7},\frac{4}{7})\), which are also applied in our loss function, accounting for the inherent feature ambiguity from the unseen directions. We provide further details in the supplementary materials.

\subsection{Evaluation Datasets}

\noindent\textbf{Mixamo Character Dataset. }
We have created a challenging dataset by gathering publicly available character meshes from the internet, using Mixamo \cite{D2} as our source. The dataset comprises a variety of characters with animations, including monsters, zombies, and characters wearing different clothes, accessories, and weapons. We have selected 100 characters with 500 different poses, and we have split them into 90 for the training set and 10 for the test set. Instead of using photorealistic renderings, which is a common approach for training frameworks, we have deliberately scaled down the quality of our renderings to replicate the in-the-wild drawings or paintings of animation characters, which contain less visual information than human photos taken from real environments. For each pose, we have rendered six images, rotating around the yaw axis with a \(60^{\circ}\) interval, resulting in a total of 3,000 images for each character. \\

\noindent\textbf{THuman2.0 Dataset. }
THuman2.0 \cite{D4} provides high-quality 3D meshes of 526 different human subjects with varying textures and poses. We render six images per subject with \(60^{\circ}\) interval in total of 3,156 images for evaluation.   

\subsection{Evaluations}
\noindent\textbf{Comparison. } We conducted extensive comparisons in three different criteria: 1) a self-curated 3D character dataset \cite{D2}, 2) a benchmark 3D human dataset \cite{D4}, and 3) in-the-wild images from the Internet. To compare, we used code-available template-free 3D human reconstruction methods, namely PIFu \cite{F3} and its improved versions, such as Geo-PIFu \cite{F6} and PIFuHD \cite{F4}, since template-based methods fail to recover character geometries entirely.
\vspace{-1em}
\begin{table*}
\begin{center}
\begin{tabular}{l | ccc | ccc}
& \multicolumn{3}{c|}{Mixamo Character Dataset \cite{D2}} 
& \multicolumn{3}{c}{THuman2.0 Dataset \cite{D4}} \\ 
Methods & Normal \(\downarrow\) & P2S \(\downarrow\) & Chamfer \(\downarrow\) & Normal \(\downarrow\) & P2S \(\downarrow\) & Chamfer \(\downarrow\) \\
\hline
PIFu \cite{F3} & 0.2629 & 0.0592 & 0.0513 & 0.2331 & 0.0487 & 0.0446\\
PIFuHD \cite{F4} & 0.2489 & 0.0508 & 0.0509 & 0.2226 & 0.0486 & 0.0473\\
Geo-PIFu \cite{F6} & 0.2353 & 0.0468 & 0.0432 & 0.2178 & 0.0483 & 0.0443\\ 
\hline
Ours (coarse only)  & 0.2067 & 0.0359 & 0.0397 & 0.2058 & 0.0480 & 0.0454\\
Ours (coarse-to-fine) & \textbf{0.2035} & \textbf{0.0336} & \textbf{0.0378} & \textbf{0.2034} & \textbf{0.0465} & \textbf{0.0438} \\
\end{tabular}
\caption{\textbf{Quantitative comparison.} Evaluation on Mixamo and THuman2.0 with competing 3D human reconstruction methods. 
}
\label{table:comparison_quant}
\end{center}
\vspace{-3em}
\end{table*}

\begin{figure*}
  \centering
  \includegraphics[width=\linewidth]{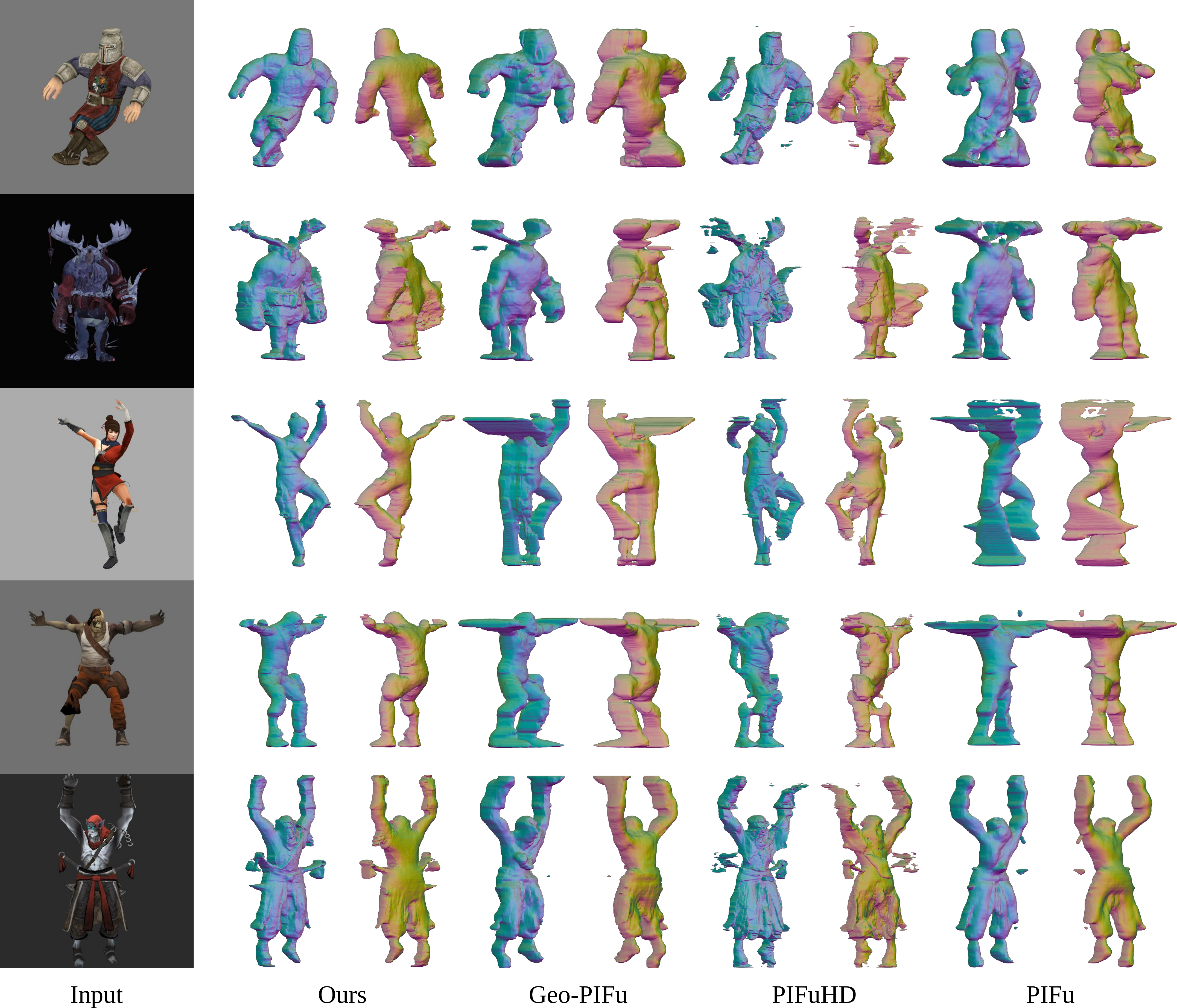}

  \caption{\textbf{Qualitative comparison.} We present results from the Mixamo dataset with varying subjects and dynamic poses.}
  \vspace{-1em}
  \label{fig:comparison}
\end{figure*}

\begin{figure*}
  \centering
  \includegraphics[width=\linewidth]{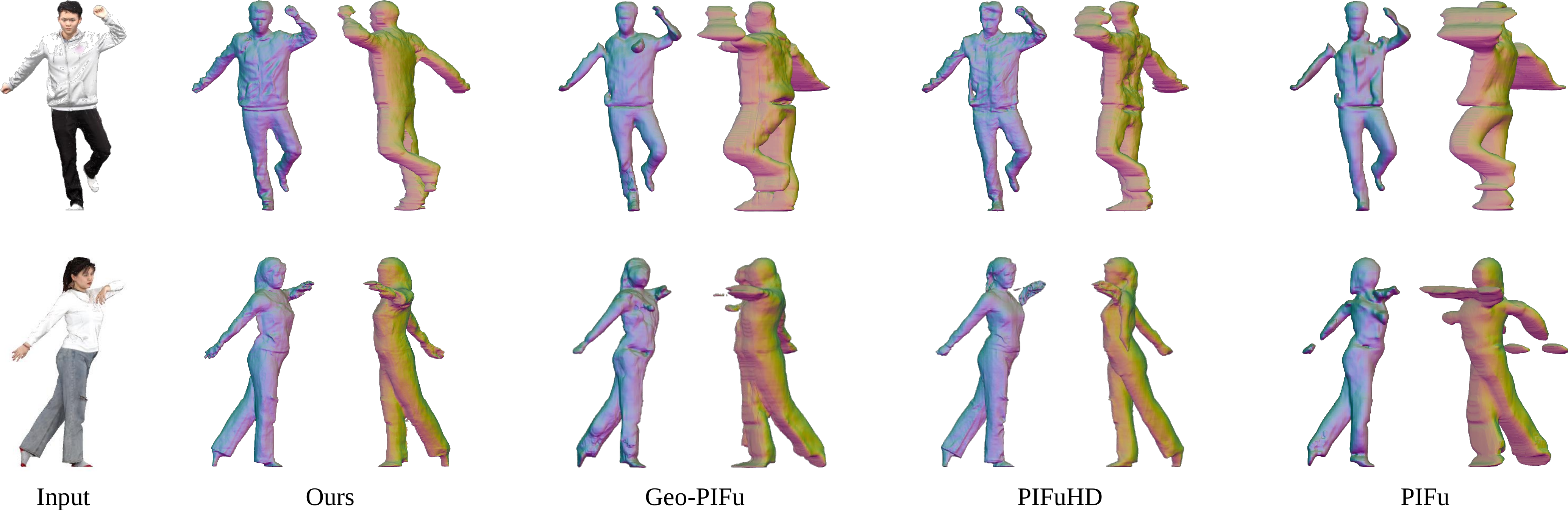}

  \caption{\textbf{Reconstructed results from a benchmark human dataset.} We show high-quality results from THuman2.0. Our results outperform prior works in capturing facial features and clothing folds with natural depictions of poses.}
  \label{fig:thuman}
\end{figure*}

\begin{figure*}
  \centering
  \includegraphics[width=\linewidth]{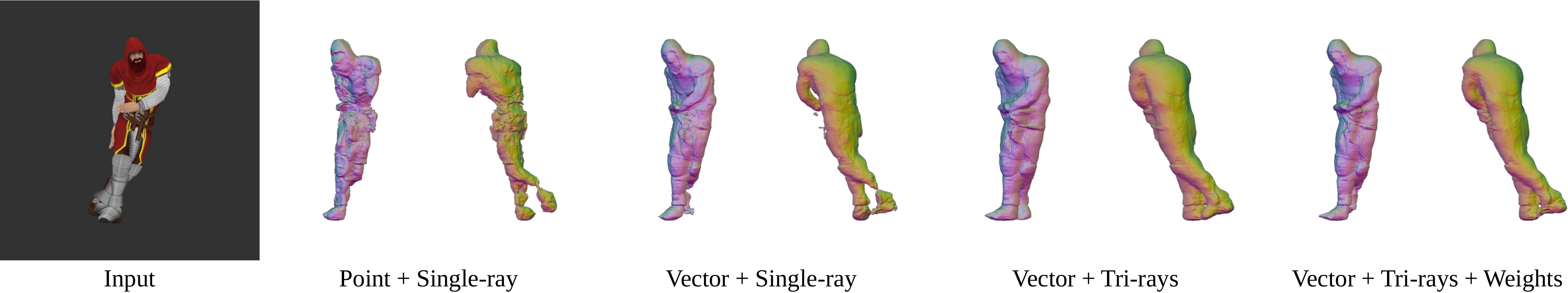}
  \caption{\textbf{Qualitative results on ablation study.} TIFu shows its advantage in reducing feature ambiguity along unseen directions. Our adaptive losses effectively mitigate the class imbalance problem in inferring mostly empty 3D space.}
  \label{fig:ablation}
\end{figure*}

\begin{figure*}
  \centering
  \includegraphics[width=\linewidth]{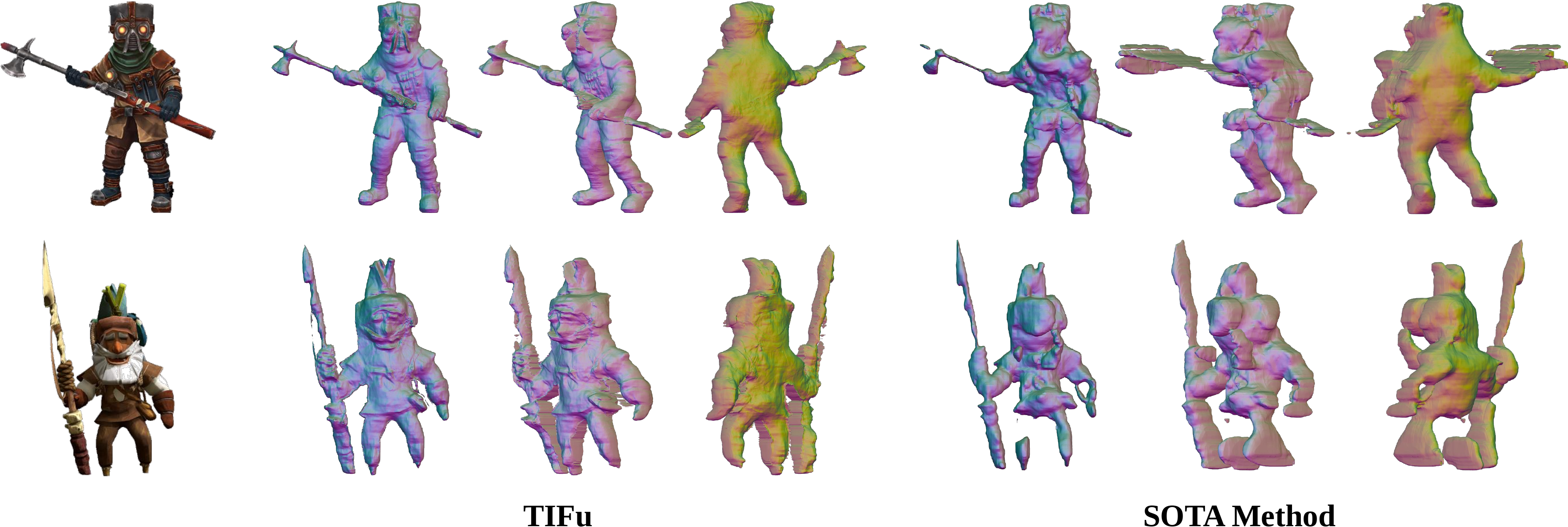}
  \caption{\textbf{Qualitative results on in-the-wild images.}}
  \label{fig:main}
  \vspace{-1em}
\end{figure*}

We reported the three most common metrics for quantitative analyses in 3D human reconstructions: normal consistency (Normal), point-to-surface (P2S) distance, and Chamfer distance. Normal calculates the L2 error between normal maps from the predicted and ground truth meshes in six different views, rotating around the yaw axis with an interval of \(60^{\circ}\). P2S and Chamfer measure the distances between two point clouds, each with 100K randomly sampled surface points from the predicted and ground truth meshes. P2S measures the one-way average distance from each point in the predicted point cloud to the closest point in the ground truth point cloud. Chamfer accounts for the two-way average distance between the two point clouds back and forth.

In Tab. \ref{table:comparison_quant} and Fig. \ref{fig:comparison}, we achieved state-of-the-art performances in all three metrics with accurate modeling of dynamic poses and fine depiction of armors, antlers, and loose clothes, among others. Geo-PIFu employed additional 3D voxel features that better captured geometric context with less dragging geometry along the depth compared to PIFu. However, both lacked high-fidelity surface details bounded by their latent feature resolution. PIFuHD was capable of detailed depiction of fine details, however, it included spiked mesh surfaces with unnaturally distorted or disconnected body parts.
With less model complexity and faster inference time compared to PIFuHD (see Tab. \ref{tab:complexity}), our approach outperformed PIFuHD in normal consistency error without surface artifacts.

We validated the generalization ability of our framework using in-the-wild images downloaded from the Internet. In Fig. \ref{fig:main}, we show reconstruction results from in-the-wild images compared to Geo-PIFu. We significantly improved details around shaded areas where a mask overlaps with garments or a nose closely placed to a mustache. Our method also reconstructs separate objects such as axes or spears coherent with the main body.
Overall, our method demonstrated superior consistency and accuracy compared to existing methods by a large margin across characters, humans, and in-the-wild images, as highlighted in Tab. \ref{table:comparison_quant}, Fig. \ref{fig:comparison}, Fig. \ref{fig:thuman}, and Fig. \ref{fig:main}.

\vspace{-1em}
\begin{table}
\scriptsize
\begin{center}
\begin{tabular}{ccc | ccc}
 \multicolumn{3}{c|}{Methods} & \multicolumn{3}{c}{Mixamo Character Dataset \cite{D2}} \\
Vector & Tri-rays & Weights & Normal \(\downarrow\) & P2S \(\downarrow\) & Chamfer \(\downarrow\) \\
\hline
 & & & 0.2430 & 0.0501 & 0.0505 \\
 \checkmark & & & 0.2109 & 0.0368 & 0.0403\\
 \checkmark & \checkmark & & 0.2083 & 0.0357 & 0.0396\\
 \checkmark & \checkmark & \checkmark & \textbf{0.2035} & \textbf{0.0336} & \textbf{0.0378}\\
\end{tabular}
\caption{\textbf{Quantitative results on ablation study.} Validation of the effectiveness in TIFu modeling shapes by vectors with adaptive weighting mask. }
\label{table:ablation_quant}
\end{center}
\vspace{-2em}
\end{table}

\begin{table}
\scriptsize
\centering
\begin{tabular}{l | cccc}

 & Ours  
 & Geo-PIFu \cite{F6}
 & PIFuHD \cite{F4} 
 & PIFu \cite{F3}\\

\hline
\# Parameters (Millions) & 109.6 & 30.6 & 387.0 & 15.6\\
Inference Time (Frames/Second) & 0.14 & 0.25 & 0.06 & 0.31\\

\end{tabular}
\caption{\small{Measured in Intel i9-7900X with NVIDIA TITAN X.}}
\label{tab:complexity}
\vspace{-2em}
\end{table}

\noindent\textbf{Ablation Study. } We evaluate the following contributions: 1) vector-level 3D shape modeling, 2) tri-directional feature learning, and 3) adaptive loss function.

In Tab. \ref{table:ablation_quant}, we present the results of our ablation study where we gradually add design choices to the baseline approach, which infers binary occupancies using a pixel-aligned implicit function. Our study compares point-level and vector-level 3D reconstruction, where the significant difference is the model's ability to capture depth-wise 3D context without distorted body parts or dragged geometry. As illustrated in Fig. \ref{fig:comparison}, previous works using point-level 3D shape modeling suffer from these issues and cannot comprehend dynamic poses and variant body sizes accurately. We found that self-occluded regions where legs are crossing each other or challenging postures where joints are bent towards the camera significantly induce PIFu-based approaches to generate false reconstructions. Previous methods are severely corrupted when inferring out-of-distribution images with higher feature ambiguity. Thus, we emphasize that TIFu modeling of the entire vector space at once is crucial for reducing depth ambiguity and recovering smoothly connected body parts.

Furthermore, tri-directional feature learning recovers more natural shapes without noisy surfaces than the single-ray approach, as depicted in Fig. \ref{fig:ablation}. Our results validate the benefit of incorporating latent shape features, including unseen directions, to mitigate the strong dependency on only pixel-level information, which is susceptible to bad local minima. Our method of leveraging latent voxel features along three orthogonal axes increased global 3D shape consistency and enabled dense 3D volume construction at an arbitrary resolution. The efficacy of modeling consistent geometries in three orthogonal directions is demonstrated by the reconstruction results viewed from side angles.

Lastly, we found that using fixed weights cannot differentiate the magnitude of imbalance in each of our vectors. Our adaptive weights improve reconstruction quality for relatively small details or thin body parts, such as facial features, hands, and cloth wrinkles, without generating excessively thick or thin surfaces.

\section{Conclusion}
\label{sec:conclusion}

We introduce a novel approach for 3D mesh recovery of animated characters from a single image, which overcomes the challenge of accurately modeling highly variant character geometries using a new 3D shape representation. 
We employ a tri-directional vector-level modeling of 3D space to reduce the inherent ambiguity in 3D reconstruction from a single image, resulting in significant improvements in generalization ability to unknown shapes, sizes, poses, and textures. 
Our method achieves state-of-the-art performance in both extensive quantitative and qualitative evaluations. 
Overall, our method represents a significant step forward in the field of single image 3D character reconstruction, opening up new opportunities for applications in areas such as animation, gaming, and virtual reality.

\bibliographystyle{splncs04}
\bibliography{egbib}
\end{document}